\pdfoutput=1
\documentclass[final,times,twocolumn,numbers]{elsarticle}

\usepackage{prletters}
\usepackage{framed,multirow,enumitem,mathrsfs,amsfonts}
\usepackage{amsmath,amssymb,mathtools,amsfonts,bm}
\usepackage[table]{xcolor}
\definecolor{lg}{gray}{0.9}
\usepackage[skins,theorems]{tcolorbox}
\tcbset{highlight math style={enhanced,
  colframe=red,colback=white,arc=0pt,boxrule=1pt}}

\usepackage{amssymb}
\usepackage{latexsym}
\usepackage{courier}
\usepackage{subcaption}
\usepackage{graphicx,array}
\usepackage{array, boldline, makecell, booktabs,amsmath}

\newcommand*\rot{\rotatebox{90}}
\usepackage{algcompatible}
\usepackage{algorithm}
\usepackage{algorithmicx}
\usepackage[noend]{algpseudocode}
\floatname{algorithm}{Procedure}
\algnewcommand{\LineComment}[1]{\State\unskip\the\therules \(\triangleright\) #1}

\newcolumntype{L}[1]{>{\raggedright\let\newline\\\arraybackslash\hspace{0pt}}m{#1}}
\newcolumntype{C}[1]{>{\centering\let\newline\\\arraybackslash\hspace{0pt}}m{#1}}
\newcolumntype{R}[1]{>{\raggedleft\let\newline\\\arraybackslash\hspace{0pt}}m{#1}}

\DeclareMathOperator*{\argmin}{arg\,min}

\makeatletter
\def\therule{\makebox[\algorithmicindent][l]{\hspace*{.5em}\vrule height 0.8\baselineskip depth .45\baselineskip}}%

\newtoks\therules
\therules={}
\def\appendto#1#2{\expandafter#1\expandafter{\the#1#2}}
\def\gobblefirst#1{
  #1\expandafter\expandafter\expandafter{\expandafter\@gobble\the#1}}%
\def\LState{\State\unskip\the\therules}
\def\pushindent{\appendto\therules\therule}%
\def\popindent{\gobblefirst\therules}%
\def\printindent{\unskip\the\therules}%
\def\printandpush{\printindent\pushindent}%
\def\popandprint{\popindent\printindent}%

\algdef{SE}[WHILE]{While}{EndWhile}[1]
  {\printandpush\algorithmicwhile\ #1\ \algorithmicdo}
  {\popandprint\algorithmicend\ \algorithmicwhile}%
\algdef{SE}[FOR]{For}{EndFor}[1]
  {\printandpush\algorithmicfor\ #1\ \algorithmicdo}
  {\popandprint\algorithmicend\ \algorithmicfor}%
\algdef{S}[FOR]{ForAll}[1]
  {\printindent\algorithmicforall\ #1\ \algorithmicdo}%
\algdef{SE}[LOOP]{Loop}{EndLoop}
  {\printandpush\algorithmicloop}
  {\popandprint\algorithmicend\ \algorithmicloop}%
\algdef{SE}[REPEAT]{Repeat}{Until}
  {\printandpush\algorithmicrepeat}[1]
  {\popandprint\algorithmicuntil\ #1}%
\algdef{SE}[IF]{If}{EndIf}[1]
  {\printandpush\algorithmicif\ #1\ \algorithmicthen}
  {\popandprint\algorithmicend\ \algorithmicif}%
\algdef{C}[IF]{IF}{ElsIf}[1]
  {\popandprint\pushindent\algorithmicelse\ \algorithmicif\ #1\ \algorithmicthen}%
\algdef{Ce}[ELSE]{IF}{Else}{EndIf}
  {\popandprint\pushindent\algorithmicelse}%
\algdef{SE}[PROCEDURE]{Procedure}{EndProcedure}[2]
   {\printandpush\algorithmicprocedure\ \textproc{#1}\ifthenelse{\equal{#2}{}}{}{(#2)}}%
   {\popandprint\algorithmicend\ \algorithmicprocedure}%
\algdef{SE}[FUNCTION]{Function}{EndFunction}[2]
   {\printandpush\algorithmicfunction\ \textproc{#1}\ifthenelse{\equal{#2}{}}{}{(#2)}}%
   {\popandprint\algorithmicend\ \algorithmicfunction}%
\algdef{SE}[DOWHILE]{Do}{DoWhile}{\algorithmicdo}[1]{\algorithmicwhile\ #1}%

\usepackage{url}
\usepackage{xcolor}
\definecolor{newcolor}{rgb}{.8,.349,.1}

\journal{Pattern Recognition Letters}

\begin{document}

\clearpage
\thispagestyle{empty}
\ifpreprint
  \vspace*{-1pc}
\fi

\clearpage

\ifpreprint
  \setcounter{page}{1}
\else
  \setcounter{page}{1}
\fi

\begin{frontmatter}

\title{On the detection-to-track association for online multi-object tracking}

\author[1]{Xufeng \snm{Lin}} 
\author[1]{Chang-Tsun \snm{Li}\corref{cor1}}
\cortext[cor1]{Corresponding author:}
\ead{changtsun.li@deakin.edu.au}
\author[2]{Victor \snm{Sanchez}}
\author[3]{Carsten \snm{Maple}}

\address[1]{School of Information Technology, Deakin University, Waurn Ponds Campus, Geelong, VIC 3216, Australia}
\address[2]{Department of Computer Science, The University of Warwick, Coventry, CV4 7AL, UK}
\address[3]{Warwick Manufacturing Group, The University of Warwick, Coventry, CV4 7AL, UK}

\received{1 May 2013}
\finalform{10 May 2013}
\accepted{13 May 2013}
\availableonline{15 May 2013}
\communicated{S. Sarkar}

\begin{abstract}
Driven by recent advances in object detection with deep neural networks, the tracking-by-detection paradigm has gained increasing prevalence in the research community of multi-object tracking (MOT). It has long been known that appearance information plays an essential role in the detection-to-track association, which lies at the core of the tracking-by-detection paradigm. While most existing works consider the appearance distances between the detections and the tracks, they ignore the statistical information implied by the historical appearance distance records in the tracks, which can be particularly useful when a detection has similar distances with two or more tracks. In this work, we propose a hybrid track association (HTA) algorithm that models the historical appearance distances of a track with an incremental Gaussian mixture model (IGMM) and incorporates the derived statistical information into the calculation of the detection-to-track association cost. Experimental results on three MOT benchmarks confirm that HTA effectively improves the target identification performance with a small compromise to the tracking speed. Additionally, compared to many state-of-the-art trackers, the DeepSORT tracker equipped with HTA achieves better or comparable performance in terms of the balance of tracking quality and speed. 
\end{abstract}


\end{frontmatter}


\section{Introduction}
\label{sec:introduction}
Multi-object tracking (MOT), which aims to track multiple objects of interest in video sequences, is an essential building block of a wide range of advanced applications such as video surveillance and autonomous driving. Driven by the great success of deep neural networks in object detection, the \emph{tracking-by-detection} paradigm has gained substantial attention in recent years. Such a paradigm first detects targets of interest in single video frames and then builds up tracks of targets by associating the detections. Thus, lying at the heart of tracking-by-detection based MOT is the detection-to-track association. To avoid any confusion, a 'target' in this manuscript refers to an object of interest that appears, moves and disappears in a camera's field of view, and a 'track' refers to a target's trajectory or path, which records all the information, including ID, positions and appearance features, associated with the target. The works proposed to address the data association problem can be categorized into \emph{offline} (a.k.a \emph{batch}) and \emph{online} approaches. Offline approaches \cite{berclaz2011multiple,butt2013multi,tang2017multiple} typically formulate the data association as a global optimization problem by considering the detections over a batch of frames or all frames. They are effective in overcoming issues like unreliable detections, similar appearance, and frequent occlusions, but their high computational cost and non-causal fashion of tracking make them unsuitable for time-critical applications. By contrast, the online approaches \cite{bewley2016simple,wojke2017simple} sequentially link the detections with tracks on a frame-by-frame basis by only considering the information up to the present frame, thus compared to the offline approaches, they are usually much more efficient but are more prone to association errors. 

Like in many other computer vision tasks, the accuracy-efficiency trade-off for MOT is quite profound. Higher accuracy of detection and association usually comes with a higher computational cost. Notably, many existing works tend to place little emphasis on tracking efficiency. Complicated modules and ad-hoc twists are usually devised to improve tracking accuracy. This is most likely because the performance ranking on the MOT benchmarking datasets is largely based on the tracking accuracy. However, for many practical applications, such as natural disasters monitoring and video surveillance systems, the tracking speed is essential for the success of the overall system. Another important aspect of the tracking performance that is sometimes overlooked is the accuracy and consistency of target identification. In practice, reliable identity information about the tracked targets is key for many high-level tasks such as abnormal behavior analysis and trajectory predictions.

Existing MOT methods typically perform detection-to-track association by exploiting multiple cues, including the motion and appearance information. The appearance information, which is usually represented as feature vectors produced by an appearance model, is important for accurate long-term tracking and identification. However, most existing association methods only consider the appearance feature distances calculated at the current frame and ignore the historical appearance feature distance records, which are also useful for track association. In this paper, we focus on the online track association based on appearance information and propose a hybrid track association (HTA) algorithm that enables more accurate and robust detection-to-track association with a small compromise to the tracking speed. We model the historical appearance distance records between the detections and tracks with an efficient incremental Gaussian mixture model (IGMM) \cite{pinto2015fast}. The statistical information derived from the IGMM is then used as auxiliary information for the association cost calculation based on appearance distance.

\section{Related work}
\label{sec:related_work}
Given the vast variety of studies on MOT, we will only review those that are most closely related to our work.

Bewley \emph{et al.} \cite{bewley2016simple} proposed a simple online and real-time tracking (SORT) algorithm, which only uses the Intersection over Union (IoU) distance between the predicted detections of the tracks (with a Kalman filter) and the detections in the current frame for the association. The appearance information and the potential occlusions are ignored in this framework. Therefore, it is fast and can easily reach a processing speed of 60$\sim$100 frames per second (FPS). However, due to the lack of appearance information in the association process, SORT shows poor identification performance in scenes with moderate occlusions. 

Wojke \emph{et al.} \cite{wojke2017simple} proposed a DeepSORT framework, which extends the SORT framework by introducing appearance information in the association process with a deep convolutional neural network (CNN) trained offline on large-scale image datasets. As the appearance information is typically represented by fixed-length feature vectors produced by an appearance model, e.g. the CNN-based person re-identification model in \cite{wojke2017simple}, we will use ``appearance information'' and ``appearance feature'' interchangeably hereinafter. Another import contribution of DeepSORT is the \emph{cascade matching strategy} (CMS) \cite{wojke2017simple}, which gives higher priorities to tracks that have been seen more recently. The authors argued that CMS is effective in reducing the incorrect association due to temporary and long-term occlusions. In fact, many earlier works \cite{kuo2010multi,yang2012online,bae2014robust,yu2016poi} have pointed out the importance of appearance information in the detection-to-track association. However, few studies have been dedicated to investigating the effects of different appearance-based association methods on tracking performance. 

Recently, Wang \emph{et al.} \cite{wang2019towards} proposed to handle the detection-to-track association based on the distances between the appearance features of the detections and the `smoothed' feature of each track, which is an exponential moving average of the appearance features of temporally adjacent detections in the same track. This strategy considers the temporal relationship of detections and thus is expected to be more accurate than CMS. More recently, Han \emph{et al.} \cite{han2020complementary} proposed to exploit the complementary information of a pair of synchronized videos captured, respectively, from a top view in a high altitude, e.g. by a drone-mounted camera, and from a horizontal view, e.g. by a helmet-mounted camera, to improve the accuracy of track association for MOT. However, it requires that both the horizontal- and top-view videos are available and synchronized prior to tracking, which may not be applicable in some practical scenarios.

Unlike the aforementioned methods that only consider the distance between appearance features at the current frame, in this work we propose a hybrid track association algorithm that also incorporates the statistical information of the historical appearance distance records. Our work bears some similarity to the association method based on tracklet confidence \cite{bae2014robust}. In that, the average of the pairwise historical appearance distances of a track is used to measure the track quality (or confidence). Our method is essentially different from \cite{bae2014robust} in twofold. Firstly, we use an IGMM to efficiently estimate the distribution, rather than the average score in \cite{bae2014robust}, of the historical distances. Secondly, the estimated IGMM is used to provide extra information for the calculation of association cost, while in \cite{bae2014robust} the average score is used to determine which tracks will be associated preferentially.

\section{Hybrid track association}
\label{sec:hta}
\subsection{Motivation}
\label{subsec:motivation}
We denote the $\ell$-dimensional appearance feature of a detection $i$ appearing at frame $t$ as $\bm{f}_{t}^{i}$ and the historical appearance features of track $j$ up to frame $t$ as $T_{t}^{j}{=}\{\bm{f}_s^j\}$, where $s$ denotes the frame indices where the target corresponding to track $j$ has appeared. To associate the detections at frame $t$ with existing tracks, the appearance feature of each detection is compared with the appearance features of each track. This results in a cost matrix with each element $d_{t}^{i,j}{=}\mathcal{F}(\bm{f}_{t}^{i}, T_{t}^{j})$ being the association cost of assigning detection $i$ to track $j$, where $\mathcal{F}$ is a function for measuring the appearance distance between a detection and a track. For instance, if we use the minimum cosine distance, then $\mathcal{F}(\bm{f}_{t}^{i}, T_{t}^{j}){=}\text{min}_{s}\;\left(1- \frac{\bm{f}_{t}^{i}\;\cdot \;\bm{f}_s^j}{\lVert\bm{f}_{t}^{i}\rVert_2\;\lVert \bm{f}_s^j\rVert_2}\right)$. With the cost matrix, the detection-to-track association can be modeled as an assignment problem and solved using the Hungarian algorithm. To reduce the risk of assigning false detections or newly emerged objects to existing tracks, a permissible maximum distance $d_{max}$ is used to ignore any association with a cost exceeding $d_{max}$.

\begin{figure}
\centerline{\includegraphics[width=0.95\columnwidth]{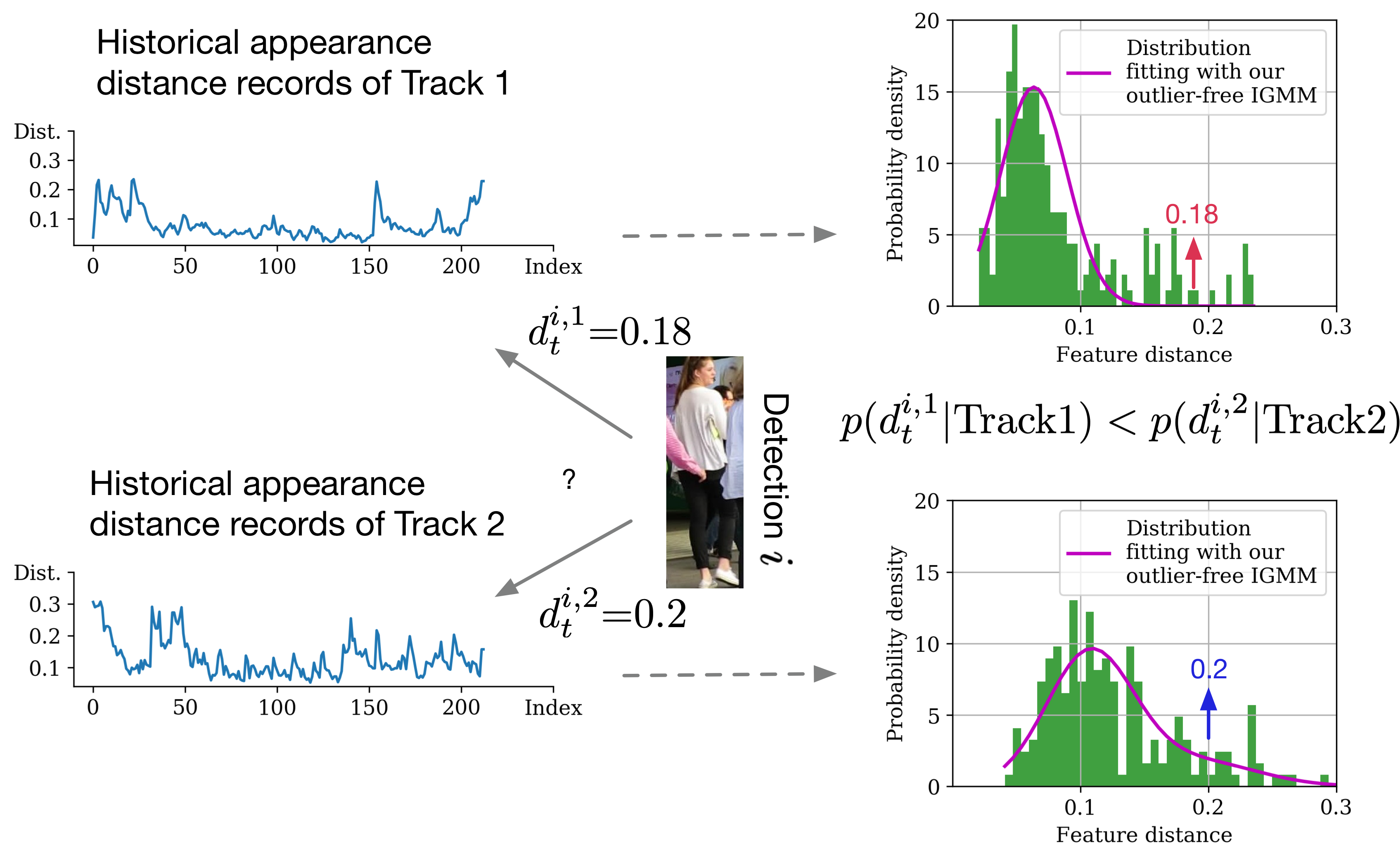}}
\caption{Exploiting the statistical information of the historical appearance distance records of a track for the detection-to-track association. The two plots on the left side, respectively, show the historical distance records for Track 1 and Track 2. The corresponding histograms are shown on the right. The distance-based assignment will wrongly assign detection $i$ to Track 1 as $d_{t}^{i,1}<d_{t}^{i,2}$, while looking at the two histograms, we can see that the likelihood of $d_{t}^{i,2}$ belonging to the distance distribution of Track 2 is higher than that of $d_{t}^{i,1}$ belonging to the distance distribution of Track 1, i.e. $p(d_{t}^{i,1}|\text{Track} 1)<p(d_{t}^{i,2}|\text{Track} 2)$. Such extra information provides a chance to correct the wrong assignment.}
\label{fig:association}
\end{figure}

The above distance-based association only considers the appearance distances at the current frame $t$. However, the historical appearance distance records may also provide useful information for reducing the ambiguities in the detection-to-track association. An example is shown in Fig. \ref{fig:association}, where the ground-truth is that detection $i$ belongs to Track 2 at frame $t$. However, due to the outliers appearing in the tracks or the limited discriminative power of the appearance features, detection $i$ has a smaller distance to Track 1 than to Track 2, i.e. $d_{t}^{i,1}<d_{t}^{i,2}$. Consequently, the above distance-based assignment will wrongly associate detection $i$ with Track 1 rather than Track 2. Meanwhile, if we look at the corresponding histograms of the historical distance records of the two tracks, it becomes evident that the likelihood of $d_{t}^{i,2}$ belonging to the distance distribution of Track 2 is higher than that of $d_{t}^{i,1}$ belonging to the distance distribution of Track 1, i.e. $p(d_{t}^{i,1}|\text{Track} 1)<p(d_{t}^{i,2}|\text{Track} 2)$, which implies that detection $i$ is more likely to belong to Track 2. Such statistical information can be exploited to correct the inaccurate associations when multiple tracks with marginally different distances are competing for the same detection. This motivates us to model the historical distance records and use the derived statistical information to aid the detection-to-track association for MOT. To achieve this purpose, when a detection $i$ is assigned to a track $j$ at a specific frame $t$, we store not only the appearance feature $\bm{f}_{t}^{i}$ but also the corresponding appearance feature distance $d^{i,j}_t$. As track $j$ builds up, this will generate the historical appearance distance records $D^j=\{d^{i,j}_t\}$, where $t$ is the frame indices where detection $i$ is found to be matched with track $j$. For brevity, we will drop the superscripts $i$ and $j$ in the following analysis and denote the historical appearance distance records of a track by $D=\{d_n\}$, where $n=1,2,...$ is the internal index of the appearance distance records in $D$.

\subsection{Incremental Gaussian mixture model}
\label{subsec:igmm}
Before delving into the details of modeling the historical distance records of a track, let us turn our attention to its relationship with foreground objects detection in video surveillance. Detecting foreground objects is usually accomplished by establishing a statistical background model at each pixel location and the pixel values that do not fit the model are considered as `foreground'. Similarly, if we consider the historical appearance distances of a track as pixel `values' appearing at a `track location' over time, modeling the historical distances of a track is equivalent to modeling the background pixel values at a pixel location. For this reason, we adopt the Gaussian mixture model, which is widely used for background modeling, to model the historical appearance distances of a track. Considering the time-series nature of the online association and the real-time requirement of tracking systems, we develop our algorithm based on a fast incremental Gaussian mixture model (IGMM) \cite{pinto2015fast}. 

\begin{figure}[t]
\begin{subfigure}{.25\textwidth}
  \centering
  \includegraphics[width=.98\linewidth]{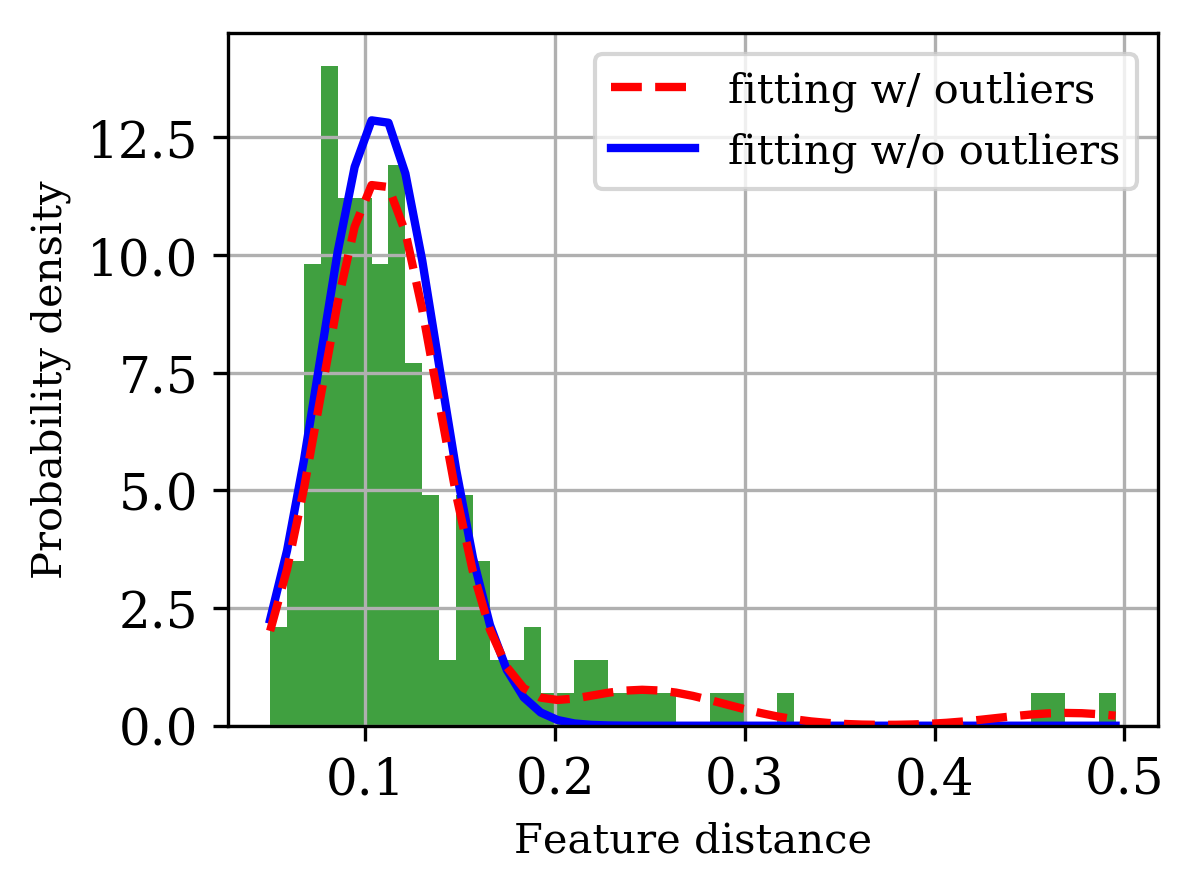}  
  \label{fig:sub-first}
\end{subfigure}%
\begin{subfigure}{.25\textwidth}
  \centering
  \includegraphics[width=.98\linewidth]{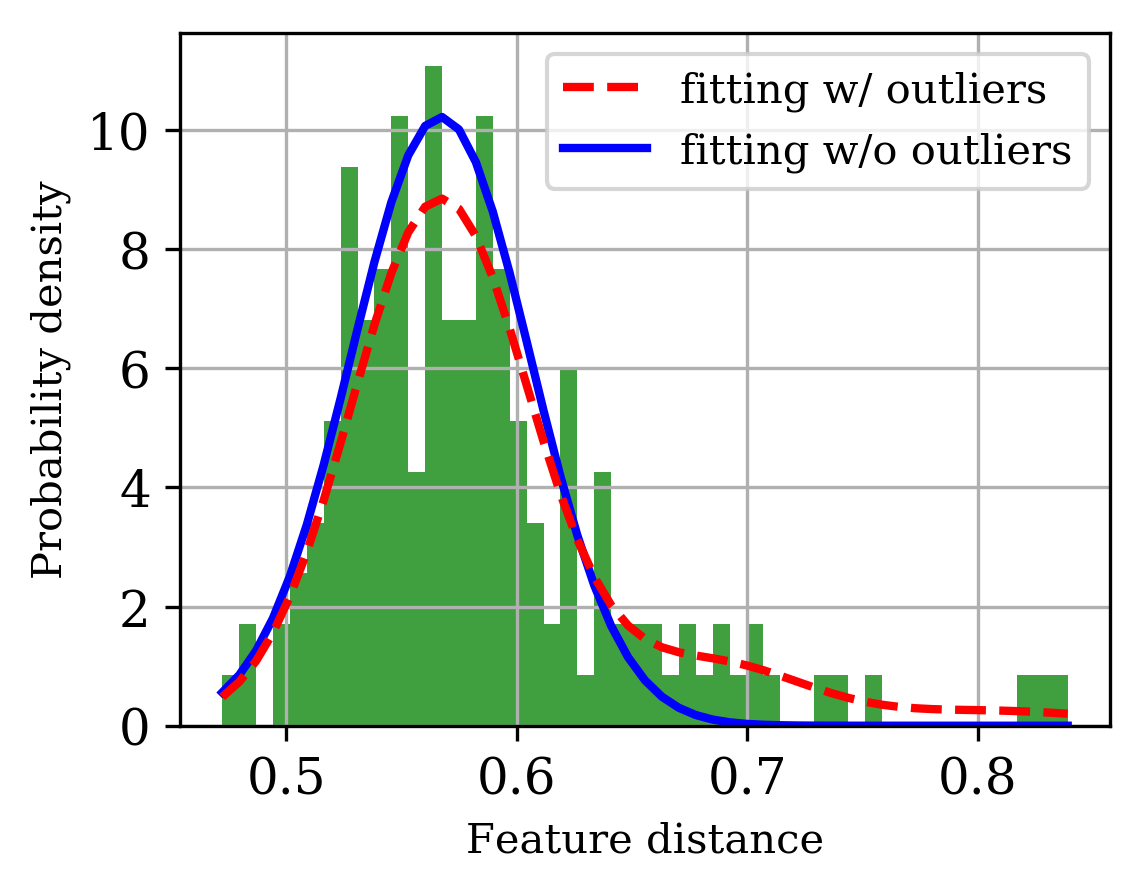}  
  \label{fig:sub-second}
\end{subfigure}
\caption{Fitting a chi-square distribution (left) and the fourth root of a chi-square distribution (right) with an IGMM. The dashed red and solid blue curves show the fitting results for all data and data without ‘outliers’, respectively.}
\label{fig:fitting}
\vspace{-5pt}
\end{figure}

Following the work in \cite{huang2020sqe}, we model the cosine distance or squared Euclidean distance\footnote{It is easy to show that they are equivalent in the case of unit-norm vectors.} between two $\ell$-dimensional unit-norm appearance features as a chi-square distribution with $\ell$ degrees of freedom (see the histograms in Fig. \ref{fig:association}). However, it is unsuitable to fit a non-norm chi-square distribution with an IGMM, so we take the fourth root of the chi-square random variable to approximately transform the chi-square distribution to a normal distribution, as suggested in \cite{hawkins1986note}. As we can see in Fig. \ref{fig:fitting}, transforming to a normal distribution allows for a more accurate fitting with IGMM. Suppose that we have the historical distance records $D=\{d_n\}$ for a track, where $n$ is the internal index of the data points in $D$ and $d_n$ is the fourth root of the distance between the appearance features of a track and a detection assigned to it. To model the distribution of the data points in $D$, we use the following Gaussian mixture model with $K_n$ components:
\begin{align}
p\left(d_n\big|\Theta_n\right)={\textstyle\sum}_{k=1}^{K_n}\pi_{k,n}p\left(d_n \big| \mu_{k,n}, \sigma^2_{k,n}\right),
\end{align} 
where $\pi_{k,n}, {\textstyle\sum}_{k=1}^{K_n}\pi_{k,n}=1$ is the mixture weight for the $k^\text{th}$ component when the $n^\text{th}$ data point arrives and $p\left(d_n \big| \mu_{k,n}, \sigma^2_{k,n}\right)$ is the $k^\text{th}$ component defined as a Gaussian distribution with mean $\mu_{k,n}$ and variance $\sigma^2_{k,n}$:
\begin{equation}
p\left(d_n\big|\mu_{k,n}, \sigma^2_{k,n}\right)=\frac{1}{\sqrt{2\pi_{k,n}\sigma^2_{k,n}}}\text{exp}\left({-\frac{\left(d_n-\mu_{k,n}\right)^2}{2\sigma^2_{k,n}}}\right).
\end{equation}
According to the Bayes' rule, the posterior probability of $d_n$ belonging to the $k^\text{th}$ component can be computed as
\begin{align}
p\left(z_k=1\big|d_n\right)=\frac{\pi_{k,n}p\left(d_n\big|\mu_{k,n}, \sigma^2_{k,n}\right)}{\sum_{m=1}^{K_n}{\pi_{m,n}p\left(d_n\big|\mu_{m,n}, \sigma^2_{m,n}\right)}},
\end{align}
where $\bm{z}=(z_1, ..., z_{K_n})$ is a vector of $K_n$ mutually exclusive binary variables with $z_k=1$ indicating the $k^\text{th}$ component is responsible for generating the data point. 

When a new data point $d_n$ arrives, i.e. a new detection is assigned to a track, the incremental estimation process is carried out by executing either Procedure 1: creating a new component or Procedure 2: updating the existing component(s), followed by an additional Procedure 3) removing spurious components.

\vspace{5pt}
\noindent{\textbf{Procedure 1: creating a new component}}
\vspace{3pt}

If there are no existing components (i.e. $K_n=0$ before the first data point arrives) or the update criterion in Procedure 2 is not met, a new component $k=K_n+1$ is created and initialized: 
\begin{align}
\begin{dcases}
v_{k,n+1}=1, N_{k,n+1}=1, \mu_{k,n+1}=d_n,  \\
K_{n+1}\leftarrow K_n+1, \sigma^2_{k,n+1}=\sigma^2_{ini}, \pi_{k,n+1}=1/{{\textstyle\sum}_{m=1}^{K_{n+1}}{N_{m,n+1}}},
\end{dcases}
\label{eqn:create}
\end{align}
where $N_{k,n}$ is the accumulated posterior probability of the $k^\text{th}$ component when the $n^\text{th}$ data point arrives, $v_{k,n}$ is a counter used for determining whether the $k^\text{th}$ component is spurious or not (see Procedure 3 below), and $\sigma^2_{ini}$ is the user-specified initial variance of a component depending on the nature of the appearance feature. 

\vspace{5pt}
\noindent{\textbf{Procedure 2: updating existing component(s)}}
\vspace{3pt}

When the squared Mahalanobis distance between a new data point $d_n$ and any component $k$ ($1\leq k \leq K_n$) is smaller than $\chi^2_{1,1-\tau}$, i.e. the $1-\tau$ percentile of a chi-square distribution with 1 degree of freedom, we update the existing components with $d_n$ rather than creating a new component: 
\begin{align}
\begin{dcases}
v_{k,n+1}=&v_{k,n} + 1 \\
N_{k,n+1}=&N_{k,n}+p\left(z_k=1\big|d_n\right) \\ 
\mu_{k,n+1}=&\mu_{k,n}+\xi_{k,n}\left(d_n-\mu_{k,n}\right) \\ 
\sigma^2_{k,n+1}=&\sigma^2_{k,n}-\xi_{k,n}\left(\sigma^2_{k,n}-\left(d_n-\mu_{k,n+1}\right)^2\right)-\omega^2_{k,n}\left(d_n-\mu_{k,n}\right)^2 \\ 
\pi_{k,n+1}=&N_{k,n+1}/{{\textstyle\sum}_{m=1}^{K_n}{N_{m,n+1}}}, 
\end{dcases}
\label{eqn:update}
\end{align}
where $\xi_{k,n}$ is a coefficient accounting for the contribution of $d_n$ to the update of the $k^\text{th}$ component:
\begin{align}
\xi_{k,n}=&\frac{p\left(z_k=1\big|d_n\right)}{N_{k,n+1}}.
\end{align}
The interfering outliers (e.g. due to occlusions and incorrect detections as shown in Fig. \ref{fig:outliers}) may also form additional components with small mixture weights. If we increase the number of components without limitation, these small components may eventually overwhelm the importance of the components corresponding to the ground truth detections. Thus, similarly to the work in \cite{stauffer1999adaptive}, we set a maximum number of components $K_{max}=5$ and discard the component with the smallest $\pi_{k,n}$ when $K_{max}$ is reached.  

\vspace{5pt}
\noindent{\textbf{Procedure 3: removing spurious components}}
\vspace{3pt}

A component $k$ is considered spurious and removed whenever $v_{k,n}>v_{min}$ and $N_{k,n}<N_{min}$. These two parameters suggest that: if at least $v_{min}$ data points have been recorded since component $k$ was created, i.e. $v_{k,n}>v_{min}$, a sufficiently significant change in component $k$ in terms of the accumulated posterior probability still cannot be observed, i.e. $N_{k,n}<N_{min}$, then component $k$ is deemed to be spurious and should be removed. As in \cite{pinto2015fast}, we set $v_{min}=5$ and $N_{min}=3$. Each of the above three procedures is followed by renormalization to ensure that ${\textstyle\sum}_{m=1}^{K_{n+1}}\pi_{m,n+1}=1$.

\subsection{Track association}
\label{subsec:track_assoc}
\begin{figure}
\centerline{\includegraphics[width=0.8\columnwidth]{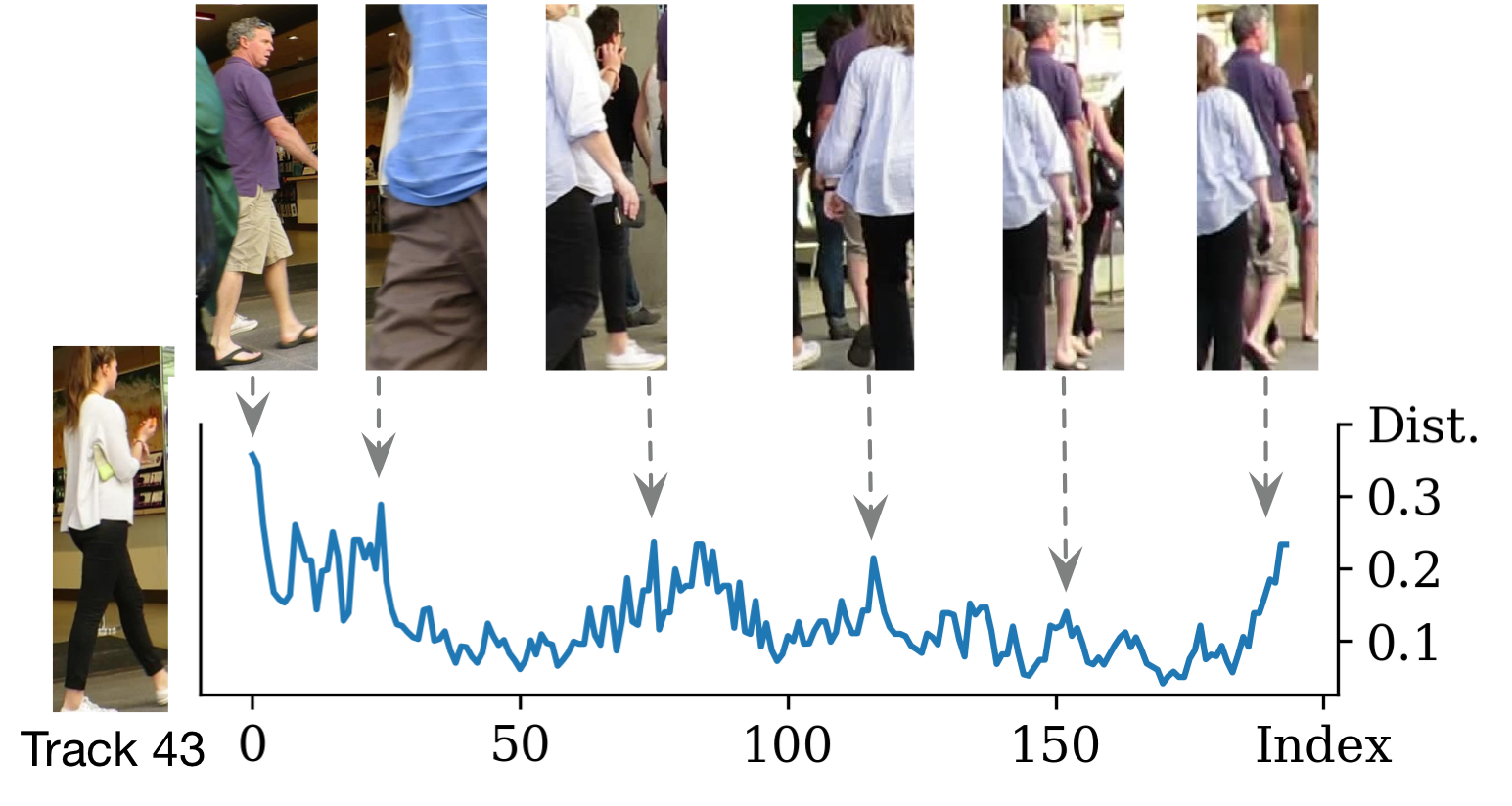}}
\caption{Examples of outliers appearing in Track 43 of sequence MOT16-09 in the MOT16 benchmark. These outliers are typically caused by occlusions or inaccurate detections, which result in relatively higher values of $d_n$.}
\label{fig:outliers}
\vspace{-5pt}
\end{figure}

As some outliers are inevitably introduced to a track due to occlusions or inaccurate detections (see the examples in Fig. \ref{fig:outliers}), we need to determine which components are most likely to be generated by the ground truth detections in the track. Heuristically, a detection belonging to the track tends to have a smaller appearance distance. Therefore, we sort the components of the estimate mixture model by the mean $\mu_{k,n}$ in descending order and choose the first $M$ components that account for at least a portion $\Upsilon$ of the model:
\begin{align}
M={\textstyle\argmin}_m\left({\textstyle\sum}_{k=1}^{m}\pi_{k,n}>\Upsilon\right),
\end{align}
where $\Upsilon\in[0, 1]$ is the minimum portion of the data that should be accounted for the ground truth detections of a track. In other words, at most $1-\Upsilon$ portion of the data is considered as `outliers'. A small $\Upsilon$ usually results in a unimodal IGMM, while a high $\Upsilon$ could results in a multi-modal IGMM, which allows for modeling more dynamic distributions, e.g. caused by scenes where tracked targets change pose constantly. We empirically set $\Upsilon{=}0.8$ in our experiments. An example is shown in Fig. \ref{fig:fitting}, where the dashed red and solid blue curves show the fitting results for all data and data without `outliers', respectively.     

With the above IGMM incrementally estimated for each track, the remaining questions are when and how to integrate such `statistical' information with the distance-only based track association method. Like many statistical models, the IGMM model will only be statistically reliable when sufficient data has been observed. For this reason, we set a minimum track length $\mathcal{L}{=}15$ to ensure that an IGMM will be estimated only for the tracks with a length no less than $\mathcal{L}$. For the tracks with a length less than $\mathcal{L}$, we only use the appearance distance as the association cost. Otherwise, we use the following hybrid association cost, hence the name \emph{Hybrid Track Association} (HTA): 
\begin{equation}
C_t= \lambda d_t + (1-\lambda)\tcbhighmath[boxrule=1pt,arc=1pt,colback=blue!10!white,colframe=black]{ \frac{\sum_{k=1}^{M}\pi_{k,n}\int_{-\infty}^{d_t}p\left(x \big| \mu_{k,n}, \sigma^2_{k,n}\right)dx}{\sum_{k=1}^{M}\pi_{k,n}}},
\label{eqn:hyrid}
\end{equation}
where $d_t$ is the distance-based term representing the appearance distance between a detection and a track at the current frame $t$. The probability-based term in the box is the cumulative probability of the outlier-free IGMM estimated with the appearance distance records observed for the track. $\lambda$ acts as a weighting factor to balance the importance of the distance-based and the probability-based terms. As can be expected, the probability-based term is not as discriminative as the distance-based term because the IGMM is only the statistical summary of the historical distances. We thus use a large weighting factor $\lambda=0.9$.

\section{Experiments}
\label{sec:experiments}

\subsection{Implementation details}
\label{subsec:details}
Our implementation is based on the DeepSORT tracker \cite{wojke2017simple}, which is mainly composed of a pedestrian detector and a track association module relying on the information provided by an appearance model and a motion estimation model based on Kalman filtering. Our proposed HTA algorithm resides in the track association module and exploits the statistical information of historical appearance distances of a track. We use YOLOv4 \cite{bochkovskiy2020yolov4} for pedestrian detection. To make the detector more generalizable, we train it on a large-scale dataset (109,471 images in total) consisting of 4 public pedestrian datasets: the Caltech dataset \cite{dollar2009pedestrian}, CUHK-SYSU dataset \cite{xiao2016end}, PRW dataset \cite{zheng2017person} and CrowdHuman dataset \cite{shao2018crowdhuman}. For the appearance model, we adopt the CNN architecture proposed in \cite{wojke2017simple} and train it on the MARS person re-identification dataset \cite{zheng2016mars}, which contains 1,067,516 bounding boxes of 1,261 pedestrians. The appearance model outputs a 512-dimensional feature for each detected bounding box. All the experiments were conducted on a PC with an NVIDIA GeForce RTX2080 Ti GPU (11 GB VRAM), Intel Core i7-8086K CPU (6 cores, 4.0 GHz), and 32 GB RAM.

\subsection{Evaluation metrics}
\label{subsec:metrics}
To evaluate the performance of MOT, we adopt the following metrics as proposed in \cite{milan2016mot16} and \cite{ristani2016performance}:
\begin{itemize}[leftmargin=*]
\setlength\itemsep{-0.05em}
\item Identification F1 score (IDF1$\uparrow$): The harmonic mean of identification precision (the ratio of the computed detections that are correctly identified) and recall (the ratio of ground truth detections that are correctly identified).
\item Multiple Object Tracking Accuracy (MOTA$\uparrow$): Overall tracking accuracy computed by combining three sources of errors: the number of false positive detections (FP$\downarrow$), the number of false negative (i.e. missed) detections (FN$\downarrow$), and the number of identity switches (IDS$\downarrow$).
\item Multiple Object Tracking Precision (MOTP$\uparrow$): Average intersection over union between the true positive detections and their corresponding ground truth targets.
\item Mostly Tracked (MT$\uparrow$):
Percentage of the ground-truth targets correctly tracked for at least 80\% of their life span.
\item Mostly Lost (ML$\downarrow$): 
Percentage of the ground-truth targets correctly tracked for at most 20\% of their life span. 
\item Fragmentations (Frag$\downarrow$): Number of times a ground truth track changes its status from `tracked' to `untracked'.
\end{itemize}
Here, `$\uparrow$' and `$\downarrow$' respectively represent that higher and lower values are preferred. IDF1 and MOTA have been widely accepted as the two most important gauges of tracking performance. They measure different aspects of a tracker: IDF1 emphasizes the accuracy and consistency of target identification, while MOTA is closely related to the detection performance. 
\begin{table*}[!ht]
\centering
\caption{Performance comparison of four different track association methods on the \texttt{training} sequences of 2D MOT15, MOT16, and MOT17. The best results for each metric are highlighted in bold. The frames per second (FPS) in the last column measures the speed of the entire algorithm including detection and association.}
  \begin{tabular}{c|l|>{\centering\arraybackslash}p{8mm}>{\centering\arraybackslash}p{10mm}>{\centering\arraybackslash}p{10mm}>{\centering\arraybackslash}p{8mm}>{\centering\arraybackslash}p{8mm}>{\centering\arraybackslash}p{8mm}>{\centering\arraybackslash}p{8mm}>{\centering\arraybackslash}>{\centering\arraybackslash}p{10mm}>{\centering\arraybackslash}p{10mm}>{\centering\arraybackslash}p{8mm}}
    \Xhline{1pt}
     &Tracker             & IDF1$\uparrow$ & MOTA$\uparrow$ & MOTP$\uparrow$ & MT$\uparrow$ & ML$\downarrow$ & FP$\downarrow$ & FN$\downarrow$ & IDS$\downarrow$ & Frag$\downarrow$ & FPS$\uparrow$ \\ \cline{1-12}
    \multirow{4}{*}{\rot{MOT15}} 
    &CMS \cite{wojke2017simple}              & 59.7\%      & 65.3\%      & 78.9\%          & 54.6\%          & \textbf{16.8\%} & 5282          & 9482 & 211  & 643   & $\sim$23 \\ 
    &$k$NN ($k{=}5$)                 & 63.0\%      & 65.3\%      & \textbf{79.0\%}          & \textbf{55.1\%} & 17.5\%        & 5359 & \textbf{9413} & \textbf{209}    & \textbf{641}   & $\sim$23 \\ 
    &EMA \cite{wang2019towards,zhang2020simple}                 & 64.1\%      & \textbf{65.4\%}      & \textbf{79.0\%}          & \textbf{55.1\%} & 17.0\%        & 5283 & 9418 & 215  & 642     & \textbf{$\sim$24} \\ 
    &HTA              & \textbf{67.7\%} & 65.1\%  & \textbf{79.0\%}          & 53.5\% & 17.2\%   & \textbf{5192} & 9580   & 278   & 647    & \textbf{$\sim$24} \\ \Xhline{1pt} 
     \multirow{4}{*}{\rot{MOT16}} 
    &CMS \cite{wojke2017simple}             & 57.3\%      & \textbf{54.6\%} & 76.8\% & 33.2\%      & 20.5\% & \textbf{9964}      & 39529    & \textbf{633}    & \textbf{1513}  & \textbf{$\sim$22} \\ 
    &$k$NN ($k{=}5$)                & 60.4\%      & 54.4\%       & \textbf{76.9\%}      & \textbf{35.0\%} & \textbf{20.1\%}         & 10299 & \textbf{39379} & 657      & 1528        & $\sim$21 \\ 
    &EMA \cite{wang2019towards,zhang2020simple}                 & 60.6\%      & 54.5\%      & 76.8\%      & 34.6\%      & 20.9\%      & 10235      & 39420      & \textbf{633}  & 1545  & $\sim$21 \\ 
    &HTA              & \textbf{62.7\%} & 54.3\%      & 76.8\%      & 33.3\%      & 20.3\%      & 9991      & 39735      & 647      & 1526        & $\sim$19 \\ \Xhline{1pt} 
    \multirow{4}{*}{\rot{MOT17}} 
    &CMS \cite{wojke2017simple}              & 57.0\%      & \textbf{54.6\%} & 76.8\% & 31.9\%      & 23.3\%    & \textbf{9329}      & 41068      & 641       & \textbf{1529}       & \textbf{$\sim$19} \\ 
    &$k$NN ($k{=}5$)                 & 60.2\%      & 54.4\%      & \textbf{76.9\%}     & 32.8\% & \textbf{22.9\%}  & 9658    & \textbf{40912}      & 659       & 1546       & $\sim$18 \\ 
    &EMA \cite{wang2019towards,zhang2020simple}                  & 60.4\%      & 54.4\%      & \textbf{76.9\%}     & 32.4\%      & 23.4\%      & 9585    & 40946      & \textbf{637}       & 1560  & \textbf{$\sim$19} \\ 
    &HTA              & \textbf{62.5\%} & 54.3\%  & \textbf{76.9\%}     & \textbf{33.1\%}      & 24.0\% & 9358      & 41275 & 660   & 1539    & $\sim$18 \\ \Xhline{1pt}  
  \end{tabular}
\label{tab:basic}
\end{table*}

\begin{figure}[!h]
\begin{subfigure}{.25\textwidth}
  \centering
  \includegraphics[width=.9\linewidth]{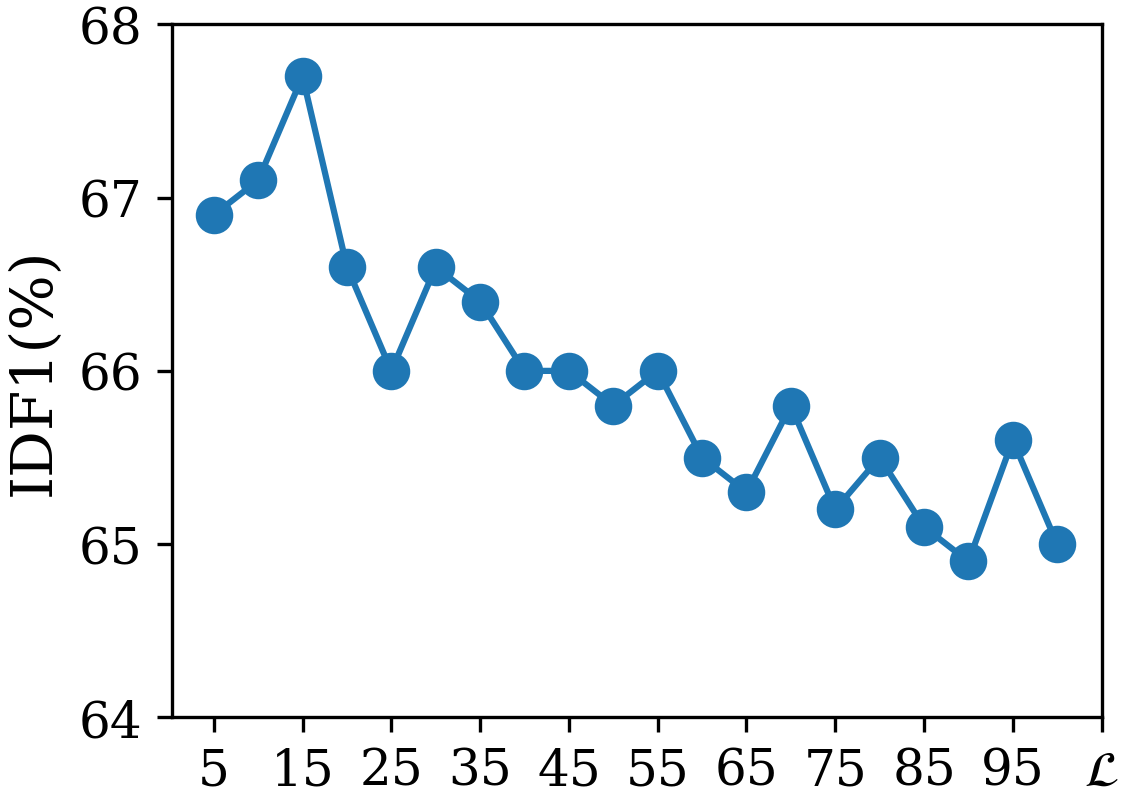}  
  \caption{}
  \label{fig:parameter-first}
\end{subfigure}%
\begin{subfigure}{.25\textwidth}
  \centering
  \includegraphics[width=.9\linewidth]{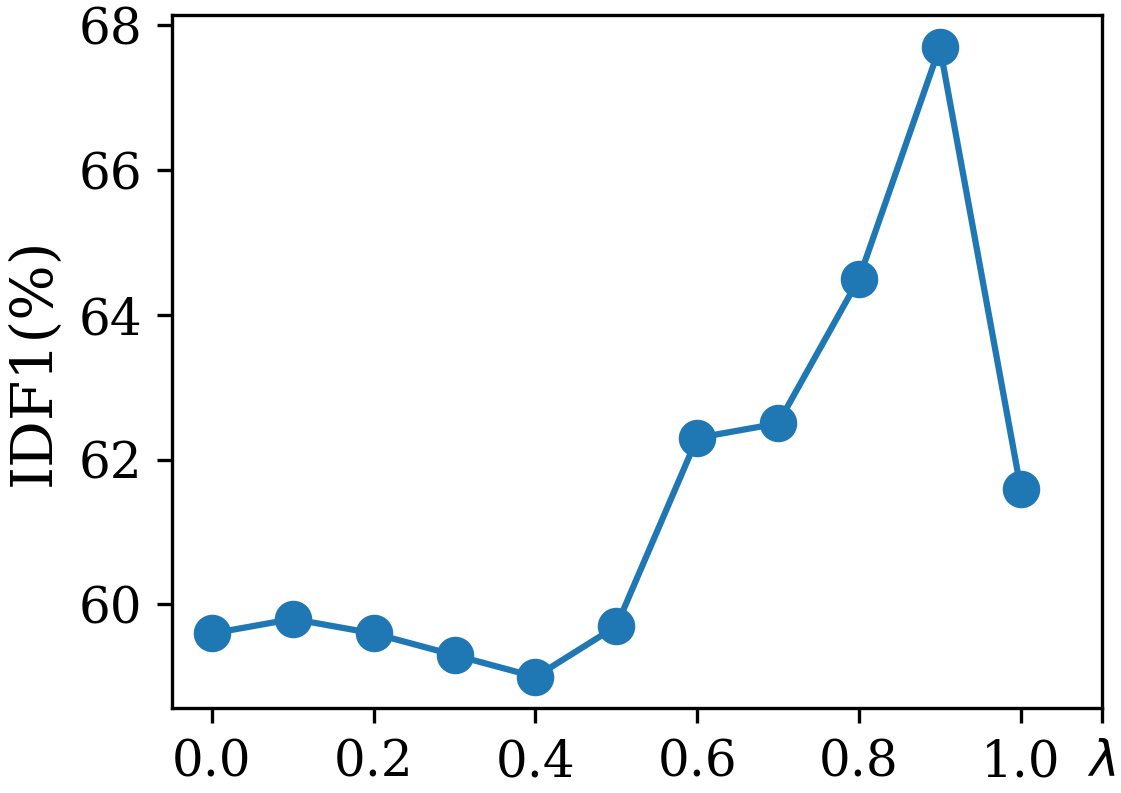}  
  \caption{}
  \label{fig:parameter-second}
\end{subfigure}
\caption{Tracking performance for various values of (a) the minimum track length $\mathcal{L}$ for reliably estimating an IGMM and (b) the weighting factor $\lambda$ (right) in Eq. (\ref{eqn:hyrid}).}
\label{fig:parameter}
\vspace{-10pt}
\end{figure}

\begin{figure}[!h]
\centering
\includegraphics[width=.9\linewidth]{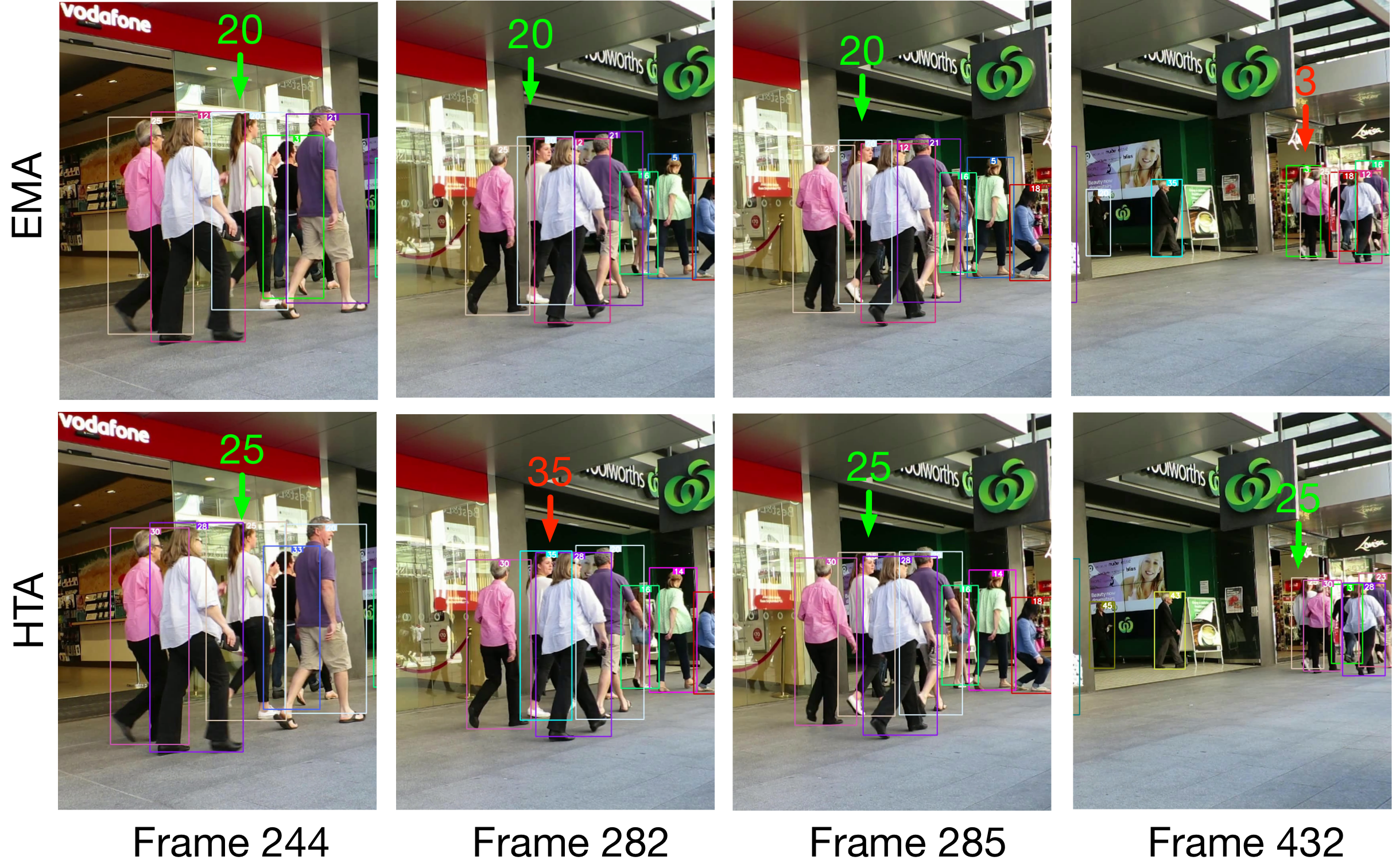}  
\caption{Qualitative evaluation on the sequence MOT16-09. This example shows how the ID of a pedestrian changes from frame 244 to frame 432 for two trackers, EMA (1st row) and HTA (2nd row). The number on each frame is the enlarged ID assigned to the pedestrian and a color change of the number indicates an ID switch. Note that for the same video sequence, different trackers may assign different IDs to the same person.}
\label{fig:qualitative_analysis}
\vspace{-10pt}
\end{figure}


\begin{table*}[!h]
\centering
\caption{Performance comparison with state-of-the-art online trackers on the \texttt{test} sequences of MOT16 and a subset of MOT15. The symbol `$\star$' represents that the tracker uses the faster-RCNN \cite{ren2015faster} based person detector provided by POI \cite{yu2016poi}, the symbol `$\S$' means that the tracker is trained on the \texttt{training} sequences of MOT16, and the symbol `$\clubsuit$' means that the results are obtained on 4 sequences of MOT15, namely PETS09-S2L1, PETS09-S2L2, ETH-Bahnhof and ETH-Sunnyday. In each column (excluding the last two rows), the bold and underlined values represent the best result of the four association methods and the best result of other methods, respectively. Except for TCODAL \cite{bae2014robust}, the frames per second (FPS) in the last column measures the speed of the entire algorithm including detection and association.}
  \begin{tabular}{l|>{\centering\arraybackslash}p{10mm}>{\centering\arraybackslash}p{10mm}>{\centering\arraybackslash}p{10mm}>{\centering\arraybackslash}p{10mm}>{\centering\arraybackslash}p{10mm}>{\centering\arraybackslash}p{10mm}>{\centering\arraybackslash}p{10mm}>{\centering\arraybackslash}>{\centering\arraybackslash}p{10mm}>{\centering\arraybackslash}p{10mm}>{\centering\arraybackslash}p{8mm}}
    \Xhline{1pt}
    Tracker     & IDF1$\uparrow$ & MOTA$\uparrow$ & MOTP$\uparrow$ & MT$\uparrow$ & ML$\downarrow$ & FP$\downarrow$ & FN$\downarrow$ & IDS$\downarrow$ & Frag$\downarrow$ & FPS$\uparrow$ \\ \cline{1-11} 

    POI$^\star$ \cite{yu2016poi}                 & 66.1\%          & 65.1\%          & 79.5\%          & 34.0\%          & 20.8\%          & 5061   & 55914               & 805           & 3093       & $<$5       \\ 
    TAP$^\star$ \cite{zhou2018online}            & \underline{73.5\%}          & 64.8\%          & 78.7\%          & 38.5\%          & 21.6\%          & 12980   & 50635                & 571           & 1048       & $<$8     \\ 
    SORT$^\star$ \cite{bewley2016simple}         & 53.8\%          & 59.8\%          & 79.6\%          & 25.4\%          & 22.7\%          & 8698    & 63245                & 1423          & 1835       & $<$12    \\ 
    VMaxx$^\star$ \cite{wan2018multi}            & 49.2\%          & 62.6\%          & 78.3\%          & 32.7\%          & 21.1\%          & 10604   &  56182               & 1389          & 1534       & $\sim$7    \\ 
    DeepSORT$^\star$ \cite{wojke2017simple}      & 62.2\%          & 61.4\%          & 79.1\%          & 32.8\%          & 18.2\%          & 12852   & 56668          & 781           & 2008       & $<$8      \\ 
    LM-CNN$^\S$$^\star$ \cite{babaee2019dual}  & 61.2\%          & 67.4\%          & 79.1\%          & 38.2\%          & 19.2\%          & 10109   & 48435          & 931           & \underline{1034}       & $\sim$2     \\ 
    EAMTT$^\S$ \cite{sanchez2016online}        & 53.3\%          & 52.5\%          & 78.8\%          & 19.0\%          & 34.9\%          & \underline{4407}    & 81223            & 910           & 1321       & $\sim$12     \\ 
    CNNMTT$^\S$ \cite{mahmoudi2019multi}       & 62.2\%          & 65.2\%          & 78.4\%          & 32.4\%          & 21.3\%          & 6578    & 55896          & 946           & 2283       & $<$6     \\ 
    JDE-1088$^\S$ \cite{wang2019towards}       & 55.8\%          & 64.4\%          & - -           & 35.4\%          & 20.0\%          & - -     & - -          & 1544          & - -        & $\sim$19     \\ 
    FairMOT$^\S$ \cite{zhang2020simple}        & 70.4\%          & \underline{68.7\%} & \underline{80.3\%} & \underline{39.5\%} & \underline{19.0\%}          & 11695   & \underline{44411}          & 953           & 2424       & \underline{$\sim$26}  \\  \Xhline{1pt} 
    CMS \cite{wojke2017simple}                                     & 58.7\%          & \textbf{62.5\%}          & 79.2\%          & 39.3\%          & 11.5\%          & 19284   & 47570          & \textbf{1456}          & \textbf{2510}       & $\sim$16 \\ 
    $k$NN ($k{=}5$)                                          & 61.3\%          & 62.3\%          & 62.3\%          & \textbf{39.4\%}          & \textbf{11.3\%}          & 19897          & \textbf{47116}          & 1692          & 2591       & $\sim$15 \\ 
    EMA \cite{wang2019towards,zhang2020simple}                                         & 61.7\%          & 62.3\%          & 79.2\%          & 39.3\%          & 11.5\%          & 19876          & 47182          & 1757          & 2635       & \textbf{$\sim$17} \\ 
    HTA                                      & \textbf{64.2\%}          & 62.4\%          & \textbf{79.3\%}          & 37.5\%          & 12.1\%          & \textbf{19071}        & 47839          & 1619          & 2529       & $\sim$15 \\ \Xhline{1pt}
    TCODAL$^\clubsuit$ \cite{bae2014robust}        & - -          & {74.3\%} & {62.9\%} & {79.8\%} & {1.5\%}          & - -   & - -          & 85           & 132       & {$<$0.5}  \\  
    HTA$^\clubsuit$                                      & {69.2\%}          & 78.4\%          & {78.5\%}          & 76.8\%          & 5.9\%          & {2607}        & 2747          & 219          & 455       & $\sim$30 \\
    \Xhline{1pt}
  \end{tabular}
\label{tab:sota}
\end{table*}

\subsection{Benchmarking datasets}
\label{subsec:benchmark}
For the evaluation of tracking performance, we use three MOT benchmark datasets: 2D MOT15 \cite{leal2015motchallenge}, MOT16 \cite{milan2016mot16}, and MOT17 \cite{milan2016mot16}. These three benchmarks, respectively, contain 22 (11 training, 11 test), 14 (7 training, 7 test), and 14 (7 training, 7 test) videos sequences in unconstrained environments filmed with both static and moving cameras. 
The ground truth annotations of the training sequences are released but those of the test sequences are unpublished to avoid over-fitting to the specific sequences \cite{milan2016mot16}. However, because the sequences in the training and test sets of these three benchmarks were captured in the same or similar environments, training on the training sets is beneficial for improving the tracking performance on the test sets. For this reason, many top-ranked trackers on the MOT benchmarks \cite{leal2015motchallenge,milan2016mot16} are trained on the training sets. It is noteworthy that we do not train on the training sets of the MOT benchmarks nor train on any dataset, e.g. the ETH dataset \cite{ess2008mobile} and PETS dataset \cite{Patino2017PETS}, that partially overlaps with the MOT benchmarks. This prevents the detector from being biased to the MOT benchmarks and allows us to conduct fair performance analysis on the training sets.

\subsection{Performance analysis}
\label{subsec:performance_analysis}

In this section, we will investigate the effect of the parameters of the proposed HTA algorithm and compare different track association methods. Experiments are conducted on the \texttt{training} sets of MOT15, MOT16, and MOT17. For MOT16 and MOT17, the detection score threshold for the detector is set to $0.3$, while for MOT15, the threshold is set to $0.7$ due to the higher number of false positives. As in \cite{wojke2017simple}, we use the cosine distance to measure the distance between appearance features and set the permissible maximum distance $d_{max}{=}0.2$. We set $\sigma^2_{ini}{=}0.005$ in Eq. (\ref{eqn:create}) for the appearance feature used in \cite{wojke2017simple}.

We create four different trackers by replacing the appearance-based association module in the DeepSORT tracker \cite{wojke2017simple} with four different track association strategies. The details of these trackers are as below:
\begin{itemize}[leftmargin=*]
\setlength\itemsep{-0.3em}
\item Cascade Matching Strategy (CMS) \cite{wojke2017simple}: This association strategy first matches the detections with the tracks that are most recently updated, and then the tracks that are second-most recently updated and so on, until all the tracks or detections are examined. The cost of associating a detection with a track is calculated as the appearance distance between the detection and the nearest detection (i.e. with the smallest distance) stored in the track.
\item $k$ Nearest Neighbors ($k$NN): This strategy associates the detections with all existing tracks at once without giving priorities to any tracks. For each detection, the association cost is the average appearance distance between the detection and its top $k$ nearest detections in each of the tracks. If the length of a track is smaller than $k$, all the detections in the track are considered. We set $k{=}5$ to strike a balance between the robustness to outliers and the accuracy of association.
\item Exponential Moving Average (EMA) \cite{wang2019towards,zhang2020simple}: This strategy performs association based on the distance between the appearance features of the detections and the `smoothed' feature of each track, which is an exponential moving average of the temporally adjacent appearance features in the same track. Unlike CMS and $k$NN, this strategy takes into consideration the temporal information and thus is expected to be more accurate than CMS and $k$NN. As in \cite{wang2019towards,zhang2020simple}, we set the weighting-decrease coefficient $\eta=0.9$.  
\item Hybrid Track Association (HTA): This strategy differs from EMA in that it considers not only the feature distance at the current frame but also the statistical information provided by the historical appearance distance records of a track.   
\end{itemize} 

We first investigate the effect of the parameters of the proposed HTA algorithm on the tracking performance. There are two important parameters for the HTA algorithm: the minimum track length $\mathcal{L}$ for reliably estimating an IGMM and the weighting factor $\lambda$ in Eq. (\ref{eqn:hyrid}). We investigate the effect of these two parameters on the IDF1 tracking performance using the \texttt{training} sets of MOT15. Fig. \ref{fig:parameter-first} and Fig. \ref{fig:parameter-second} show the results for fixed $\lambda=0.9$ and varying $\mathcal{L}\in[5, 100]$ and for fixed $\mathcal{L}=15$ and varying $\lambda\in[0, 1]$, respectively. As shown in Fig. \ref{fig:parameter-first}, the performance improves until the minimum track length $\mathcal{L}$ reaches $15$ and trends downward as $\mathcal{L}$ continues to increase. This is not surprising because integrating the `statistical' information derived from the IGMM is more beneficial in the earlier period of a track, when the appearance cues have not been sufficiently collected. In Fig. \ref{fig:parameter-second}, $\lambda=0$ and $\lambda=1$, respectively, correspond to the probability-only and distance-only associations. We can see that the peak performance is achieved at $\lambda=0.9$. We will use $\mathcal{L}=15$ and $\lambda=0.9$ for the HTA algorithm in the following experiments.

We then compare the tracking performance of different track association methods. The results on MOT15, MOT16, and MOT17 are reported in Table \ref{tab:basic}. As we can see, there is no substantial difference in MOTA for the four trackers as the MOTA is highly dependent on the detection performance. While in terms of IDF1, the proposed HTA outperforms the other three algorithms consistently, followed by EMA and $k$NN. CMS proposed in \cite{wojke2017simple} is the worst-performing tracker, with about $6\%{\sim}8\%$ lower IDF1 than HTA and $3\%{\sim}4\%$ lower IDF1 than EMA and $k$NN. Based on these results, we can gain some insights for track association: 1) CMS is not as effective as expected in improving MOTA and tends to have a negative impact on IDF1. 2) The performance gap between HTA \& EMA and CMS \& $k$NN implies that temporal information is important for accurate track association. 3) Integrating the historical information of a track is beneficial for enhancing the performance of target identification. We also note that HTA may give rise to more frequent ID switching. A qualitative analysis is shown in Fig. \ref{fig:qualitative_analysis}, where HTA makes 2 ID switches and EMA makes only 1 ID switch. Despite EMA's lower occurrences of ID switching, the wrong ID assignment at frame 432 persists until the end of the track due to the lack of an effective error correction mechanism. While for HTA, the integration of the statistical information of historical records gives a chance to quickly correct the wrong ID assignment at frame 285. As for the tracking speed, HTA and $k$NN are slightly slower than CMS and EMA, with about $1{\sim}3$ FPS slower, because extra computation is needed for estimating the IGMM or searching for the top $k$ nearest appearance features of a track.

\subsection{Comparisons with state-of-the-art trackers}
\label{subsec:sota}
We compare the aforementioned four trackers with several state-of-the-art \emph{online} trackers on the \texttt{test} sequences of MOT16 benchmark. All the algorithms are run under the private protocol, i.e. using private detectors. The comparison results are presented in Table \ref{tab:sota}. For CMS, $k$NN, EMA, and HTA, we use the same parameter settings as on the \texttt{training} sequences. The best results for these four trackers are highlighted in bold, while the best results for other trackers are underlined. We also show in the last two rows of Table \ref{tab:sota} the comparison results between HTA and the association method based on tracklet confidence and online discriminative appearance learning (TCODAL) \cite{bae2014robust} on 4 sequences of MOT15, namely PETS09-S2L1, PETS09-S2L2, ETH-Bahnhof and ETH-Sunnyday. Note that the tracking speed of TCODAL shown in Table \ref{tab:sota} does not include the time used for detection.

Compared to other state-of-the-art trackers, the proposed HTA achieves better or comparable performance in terms of the balance of tracking quality and speed. We can see that many top-performing trackers (e.g. those indicated with superscript $\star$) use the person detector provided in \cite{yu2016poi}, which is based on the faster-RCNN \cite{ren2015faster} and trained on both public and private datasets. The faster-RCNN detector delivers high-quality detections, as reflected by the high MOTA and relatively low FP and FN of POI in the second row of Table \ref{tab:sota}, but it also substantially compromises the speed of tracking. For instance, most of them can only process less than 8 frames per second, which makes them unsuitable for time-critical applications. 

The two algorithms that are faster than our proposed HTA are JDE-1088 \cite{wang2019towards} and FairMOT \cite{zhang2020simple}. Both trackers use a joint framework that shares the feature maps for the tasks of object detection and appearance feature learning, thus boosting the tracking speed. They mainly differ in the backbone network and object detection mechanism (i.e. anchor-based or anchor-free). It is noteworthy that both of JDE-1088 and FairMOT are trained on the \texttt{training} sets of MOT benchmarks. While this is beneficial for improving the performance on the \texttt{test} sets of MOT benchmarks, it may lead to overfitting to the MOT benchmarks for both detection and target identification. For instance, we have observed a large performance gap (in terms of miss detections and ID switches) for FairMOT when evaluated on videos that are quite different from the sequences in MOT benchmarks (e.g. the videos downloaded from YouTube), but the lack of tracking annotations on these datasets prevents us from providing quantitative results here. This is an issue that should not be overlooked because training on self-collected datasets may be infeasible for many practical scenarios. This is the reason why we train our pedestrian detector and appearance model on datasets that do not overlap with the MOT benchmark datasets. This enables our trackers to deliver relatively consistent performance in a plug-and-play way for unseen datasets.  

\section{Conclusions}
\label{sec:conclusions}
In this paper, we have presented a hybrid track association method that enables a more accurate and robust online detection-to-track association. Our proposed method efficiently models the historical appearance distance records of a track with an incremental Gaussian mixture model and integrates the derived statistical information into the calculation of track association cost. Evaluations on public multi-object tracking benchmarks demonstrate that, with detections provided by a real-time object detector, our proposed hybrid track association strategy achieves better or comparable performance than many other state-of-the-art online trackers in terms of the balance between tracking speed and quality.   

\section*{Acknowledgments}
This work is jointly supported by the Defence Science and Technology Laboratory (DSTL) of the Ministry of Defence of the United Kingdom through the project entitled R-DIPS: Real-time Detection of Concealment of Intent for Passenger Screening (Project No. ACC6008031) and PETRAS National Centre of Excellence for IoT Systems Cybersecurity (Project No. EP/S035362/1).

\bibliographystyle{ieeetr}
\bibliography{Manuscript}

\begin{thebibliography}{10}

\bibitem{berclaz2011multiple}
J.~Berclaz, F.~Fleuret, E.~Turetken, and P.~Fua, ``Multiple object tracking
  using k-shortest paths optimization,'' {\em IEEE Trans. Pattern Anal. Mach.
  Intell.}, vol.~33, no.~9, pp.~1806--1819, 2011.

\bibitem{butt2013multi}
A.~A. Butt and R.~T. Collins, ``Multi-target tracking by lagrangian relaxation
  to min-cost network flow,'' in {\em Proc. IEEE Comput. Soc. Conf. Comput.
  Vis. Pattern Recognit.}, pp.~1846--1853, 2013.

\bibitem{tang2017multiple}
S.~Tang, M.~Andriluka, B.~Andres, and B.~Schiele, ``Multiple people tracking by
  lifted multicut and person re-identification,'' in {\em Proc. IEEE Comput.
  Soc. Conf. Comput. Vis. Pattern Recognit.}, pp.~3539--3548, 2017.

\bibitem{bewley2016simple}
A.~Bewley, Z.~Ge, L.~Ott, F.~Ramos, and B.~Upcroft, ``Simple online and
  realtime tracking,'' in {\em Proc. IEEE Int. Conf. Image Process.},
  pp.~3464--3468, 2016.

\bibitem{wojke2017simple}
N.~Wojke, A.~Bewley, and D.~Paulus, ``Simple online and realtime tracking with
  a deep association metric,'' in {\em Proc. IEEE Int. Conf. Image Process.},
  pp.~3645--3649, 2017.

\bibitem{pinto2015fast}
R.~C. Pinto and P.~M. Engel, ``A fast incremental gaussian mixture model,''
  {\em {PloS One}}, vol.~10, no.~10, pp.~e0139931--e0139931, 2015.

\bibitem{kuo2010multi}
C.-H. Kuo, C.~Huang, and R.~Nevatia, ``Multi-target tracking by on-line learned
  discriminative appearance models,'' in {\em Proc. IEEE Comput. Soc. Conf.
  Comput. Vis. Pattern Recognit.}, pp.~685--692, 2010.

\bibitem{yang2012online}
B.~Yang and R.~Nevatia, ``Online learned discriminative part-based appearance
  models for multi-human tracking,'' in {\em Proc. Eur. Conf. Comput. Vis.},
  pp.~484--498, 2012.

\bibitem{bae2014robust}
S.-H. Bae and K.-J. Yoon, ``Robust online multi-object tracking based on
  tracklet confidence and online discriminative appearance learning,'' in {\em
  Proc. IEEE Comput. Soc. Conf. Comput. Vis. Pattern Recognit.},
  pp.~1218--1225, 2014.

\bibitem{yu2016poi}
F.~Yu, W.~Li, Q.~Li, Y.~Liu, X.~Shi, and J.~Yan, ``{POI: Multiple object
  tracking with high performance detection and appearance feature},'' in {\em
  Proc. Eur. Conf. Comput. Vis.}, pp.~36--42, 2016.

\bibitem{wang2019towards}
Z.~Wang, L.~Zheng, Y.~Liu, and S.~Wang, ``Towards real-time multi-object
  tracking,'' {\em arXiv preprint arXiv:1909.12605}, 2019.

\bibitem{han2020complementary}
R.~Han, W.~Feng, J.~Zhao, Z.~Niu, Y.~Zhang, L.~Wan, and S.~Wang,
  ``Complementary-view multiple human tracking,'' in {\em Proc. AAAI Conf.
  Artif. Intell.}, pp.~10917--10924, 2020.

\bibitem{huang2020sqe}
Y.~Huang, F.~Zhu, Z.~Zeng, X.~Qiu, Y.~Shen, and J.~Wu, ``{SQE: A self quality
  evaluation metric for parameters optimization in multi-object tracking},'' in
  {\em Proc. IEEE Comput. Soc. Conf. Comput. Vis. Pattern Recognit.},
  pp.~8306--8314, 2020.

\bibitem{hawkins1986note}
D.~M. Hawkins and R.~Wixley, ``A note on the transformation of chi-squared
  variables to normality,'' {\em The American Statistician}, vol.~40, no.~4,
  pp.~296--298, 1986.

\bibitem{stauffer1999adaptive}
C.~Stauffer and W.~E.~L. Grimson, ``Adaptive background mixture models for
  real-time tracking,'' in {\em Proc. IEEE Comput. Soc. Conf. Comput. Vis.
  Pattern Recognit.}, vol.~2, pp.~246--252, 1999.

\bibitem{bochkovskiy2020yolov4}
A.~Bochkovskiy, C.-Y. Wang, and H.-Y.~M. Liao, ``Yolov4: Optimal speed and
  accuracy of object detection,'' {\em arXiv preprint arXiv:2004.10934}, 2020.

\bibitem{dollar2009pedestrian}
P.~Doll{\'a}r, C.~Wojek, B.~Schiele, and P.~Perona, ``Pedestrian detection: A
  benchmark,'' in {\em Proc. IEEE Comput. Soc. Conf. Comput. Vis. Pattern
  Recognit.}, pp.~304--311, 2009.

\bibitem{xiao2016end}
T.~Xiao, S.~Li, B.~Wang, L.~Lin, and X.~Wang, ``End-to-end deep learning for
  person search,'' {\em arXiv preprint arXiv:1604.01850}, vol.~2, no.~2, 2016.

\bibitem{zheng2017person}
L.~Zheng, H.~Zhang, S.~Sun, M.~Chandraker, Y.~Yang, and Q.~Tian, ``Person
  re-identification in the wild,'' in {\em Proc. IEEE Comput. Soc. Conf.
  Comput. Vis. Pattern Recognit.}, pp.~1367--1376, 2017.

\bibitem{shao2018crowdhuman}
S.~Shao, Z.~Zhao, B.~Li, T.~Xiao, G.~Yu, X.~Zhang, and J.~Sun, ``Crowdhuman: A
  benchmark for detecting human in a crowd,'' {\em arXiv preprint
  arXiv:1805.00123}, 2018.

\bibitem{zheng2016mars}
L.~Zheng, Z.~Bie, Y.~Sun, J.~Wang, C.~Su, S.~Wang, and Q.~Tian, ``{MARS: A
  video benchmark for large-scale person re-identification},'' in {\em Proc.
  Eur. Conf. Comput. Vis.}, pp.~868--884, 2016.

\bibitem{milan2016mot16}
A.~Milan, L.~Leal-Taix{\'e}, I.~Reid, S.~Roth, and K.~Schindler, ``{MOT16: A
  benchmark for multi-object tracking},'' {\em arXiv preprint
  arXiv:1603.00831}, 2016.

\bibitem{ristani2016performance}
E.~Ristani, F.~Solera, R.~Zou, R.~Cucchiara, and C.~Tomasi, ``Performance
  measures and a data set for multi-target, multi-camera tracking,'' in {\em
  Proc. Eur. Conf. Comput. Vis.}, pp.~17--35, 2016.

\bibitem{zhang2020simple}
Y.~Zhang, C.~Wang, X.~Wang, W.~Zeng, and W.~Liu, ``A simple baseline for
  multi-object tracking,'' {\em arXiv preprint arXiv:2004.01888}, 2020.

\bibitem{ren2015faster}
S.~Ren, K.~He, R.~Girshick, and J.~Sun, ``{Faster R-CNN: Towards real-time
  object detection with region proposal networks},'' in {\em Proc. the Int.
  Conf. Neural Inf. Process. Syst.}, pp.~91--99, 2015.

\bibitem{zhou2018online}
Z.~Zhou, J.~Xing, M.~Zhang, and W.~Hu, ``Online multi-target tracking with
  tensor-based high-order graph matching,'' in {\em Proc. IEEE Int. Conf.
  Pattern Recognit.}, pp.~1809--1814, 2018.

\bibitem{wan2018multi}
X.~Wan, J.~Wang, Z.~Kong, Q.~Zhao, and S.~Deng, ``Multi-object tracking using
  online metric learning with long short-term memory,'' in {\em Proc. IEEE Int.
  Conf. Image Process.}, pp.~788--792, 2018.

\bibitem{babaee2019dual}
M.~Babaee, Z.~Li, and G.~Rigoll, ``{A dual CNN-RNN for multiple people
  tracking},'' {\em Neurocomputing}, vol.~368, pp.~69--83, 2019.

\bibitem{sanchez2016online}
R.~Sanchez-Matilla, F.~Poiesi, and A.~Cavallaro, ``Online multi-target tracking
  with strong and weak detections,'' in {\em Proc. Eur. Conf. Comput. Vis.},
  pp.~84--99, 2016.

\bibitem{mahmoudi2019multi}
N.~Mahmoudi, S.~M. Ahadi, and M.~Rahmati, ``{Multi-target tracking using
  CNN-based features: CNNMTT},'' {\em Multimedia Tools Appl.}, vol.~78, no.~6,
  pp.~7077--7096, 2019.

\bibitem{leal2015motchallenge}
L.~Leal-Taix{\'e}, A.~Milan, I.~Reid, S.~Roth, and K.~Schindler, ``Motchallenge
  2015: Towards a benchmark for multi-target tracking,'' {\em arXiv preprint
  arXiv:1504.01942}, 2015.

\bibitem{ess2008mobile}
A.~Ess, B.~Leibe, K.~Schindler, and L.~Van~Gool, ``A mobile vision system for
  robust multi-person tracking,'' in {\em Proc. IEEE Comput. Soc. Conf. Comput.
  Vis. Pattern Recognit.}, pp.~1--8, 2008.

\bibitem{Patino2017PETS}
L.~{Patino}, T.~{Nawaz}, T.~{Cane}, and J.~{Ferryman}, ``{PETS 2017: Dataset
  and challenge},'' in {\em Proc. IEEE Comput. Soc. Conf. Comput. Vis. Pattern
  Recognit. Workshops}, pp.~2126--2132, 2017.

\end{thebibliography}
\end{document}